# ASU-CNN: An Efficient Deep Architecture for Image Classification and Feature Visualizations

**Jamshaid Ul Rahman[1,2,*], Faiza Makhdoom[2], Dianchen Lu[1]**

[1]School of Mathematical Sciences, Jiangsu University 301 Xuefu road, Zhenjiang 212013, China.
[2]Abdus Salam School of Mathematical Sciences, GC University, Lahore 54600, Pakistan.

[*]*jamshaidrahman@gmail.com*

## Abstract

Activation functions play a decisive role in determining the capacity of Deep Neural Networks (DNNs) as they enable neural networks to capture inherent nonlinearities present in data fed to them. The prior research on activation functions primarily focused on the utility of monotonic or non-oscillatory functions, until Growing Cosine Unit (GCU) broke the taboo for a number of applications. In this paper, a Convolutional Neural Network (CNN) model named as ASU-CNN is proposed which utilizes recently designed activation function ASU across its layers. The effect of this non-monotonic and oscillatory function is inspected through feature map visualizations from different convolutional layers. The optimization of proposed network is offered by Adam with a fine-tuned adjustment of learning rate. The network achieved promising results on both training and testing data for the classification of CIFAR-10. The experimental results affirm the computational feasibility and efficacy of the proposed model for performing tasks related to the field of computer vision.

**Keywords:**

Amplifying Sine Unit, Activation Function, Convolutional Neural Networks, Oscillatory Activation, Image Classification, CIFAR-10

## 1 Introduction

Artificial Intelligence (AI) have technologically revolutionized the research and applications in different areas of science and industry [1]. Artificial Neural Networks (ANNs) [2, 3] as AI tools are replacing or more cautiously overtaking the human intelligence and duties in different aspects. ANNs are being widely used in numerical simulations [4] of problems arising in various fields of applied sciences and industrial research. In modern era, Computer vision [5, 6] is one of the most alluring area of research that deals with various tasks including image, speech and action recognition mostly for social security and safety. Most of the computer vision applications rely on the image or video-based information for recognition or classification purposes. CNNs [7] are the finest DNN architectures [8] to deal with the image-based data [9]. Due to the utility and high demand of performing different tasks, CNNs have been evolved into more robust and finer versions such as ResNets [10, 11]. The main focus of research in this area is based on the modifications of CNNs with newly devised and more efficient activation functions, loss functions and optimizers [12, 13]. The choice of loss functions mainly depends upon the nature of task to be performed, for instance, human face recognition problems can be well-posed by modifying softmax loss [14] such as Sphereface [15] and different additive parameter approaches [16, 17, 18].

Activation functions play a crucial role in determining the performance of any DNN architecture [19]. In past, different monotonic activation functions [20, 21] have been introduced but most of them faced gradient based shortcomings [22, 23] during backpropagation which were then alleviated by [24, 25] and non-monotonic activation functions such as Swish [26] and Mish [27]. All these activation functions were non-oscillatory in nature and the researchers did not pay attention to the utilization of non-oscillatory functions for the networks. In case of CNNs, ReLu has been considered as the most effective choice to be used in convolutional layers. In 2021, the introduction of GCU [28], an oscillatory non-monotonic activation function, expanded the horizon of activation functions applicable in different DNN architectures. Oscillatory activation functions [29] can perform complex assigned tasks with lesser neurons and thus are computationally cheaper than non-oscillatory activations. An oscillatory and non-monotonic activation function ASU [30] was initially proposed for DNNs simulating nonlinear dynamical systems and have potential computational favorability for nonlinear evolution problems [31], microelectromechanical systems (MEMS) [32, 33] and mechanical vibrations [34, 35]. ASU given by (1) outperformed GCU in recovering periodic and nonlinear dynamics of MEMS.

$$ASU(z) = z.\sin(z) \qquad (1)$$

In this work, we explain the integration of ASU in CNNs for image classification tasks. A simple CNN architecture has been proposed which utilizes ASU as an activation of both convolutional and dense layers. The classification is performed on publically available dataset CIFAR-10 [36] which contains 60,000 images belonging to 10 different classes. CIFAR-10 consists of 32x32 resolution colored images and there are 6000 images per class. Being a multi-class classification problem, the output layer of network is activated through softmax [37]. The dataset has non-overlapping classes of ten different easily recognizable objects, therefore sparse categorical cross-entropy loss [38, 39] is used in the network. In order to optimize the loss, Adam [40] has been used with learning rate annealing where the learning rates are decayed exponentially [41]. The implementation is being done through Tensorflow framework [42, 43].

## 2 The CNN architecture

The CNNs are considered as the most vigilant architectures to deal with the data based on images. There is a hierarchy of convolutional layers involving pooling layers followed by a dense network. Images are fed to the first convolutional layer of network where multiple features of input images are detected with the help of different filters and this process continues throughout the convolutional part of the network. Fig. 1 represents the general structure of CNN being utilized in this work for the classification of CIFAR-10. The functionality of different layers of CNN is explained in the following subsections.

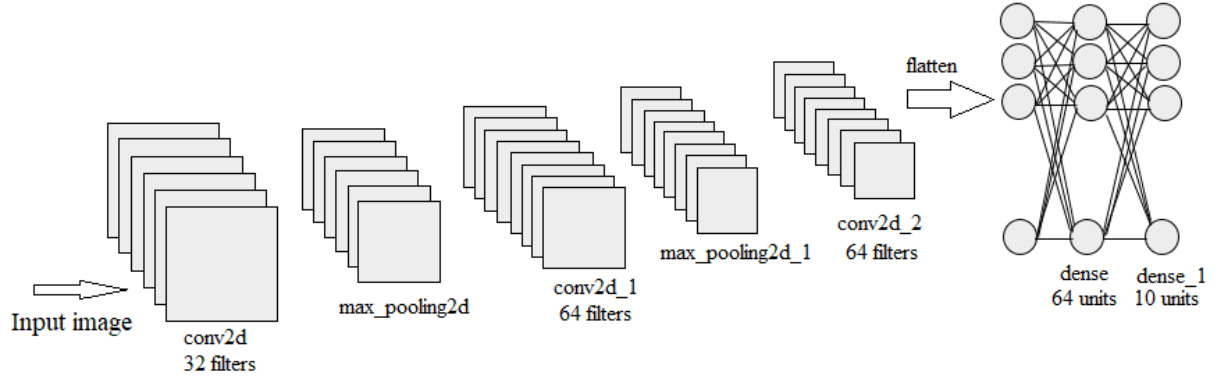

*Fig. 1* CNN having three convolutional layers with different filters to extract the features of input images and a dense network at the end for classification

**2.1 Convolutional layers**

The convolutional layer consists of a number of filters responsible for the detection of different types of features present in the input. These filters slide over the input to collect the weighted sum as in (2).

$$z^l = W^l . A^{l-1} + B^l \tag{2}$$

Where $l$ denotes specific convolutional layer of the network, $W^l$ is the matrix containing the values (weights) assigned to the filters, $A^{l-1}$ is the output (feature maps) from the previous layer and $B^l$ is bias added to the current layer. The designed architecture have three such convolutional layers i.e. $l = 1, 2, 3$ where the first one has 32 filters while second and third layers have 64 filters with valid padding and stride 1. Therefore, the dimensions of feature maps after every convolution follow (3).

$$F^l = (\left\lfloor \frac{n+2p-f}{s} + 1 \right\rfloor, \left\lfloor \frac{n+2p-f}{s} + 1 \right\rfloor, n_f) \tag{3}$$

Where $n$ represents input size, $f$ is the filter's size, $p$ stands for padding being applied, $s$ is the stride value and $n_f$ is the number of filters in that specific layer.

**2.1.1 Non-linearity layer**

This layer is responsible for adding some non-linearity in the convolved features with the help of activation functions. In this paper, the designed CNN applies ASU to (2) which is the linear output of convolution.

$$A^l = ASU^l(Z^l) \tag{4}$$

Equation (4) provides the general mathematical representation of feature maps $A^l$ being activated by ASU. Fig. 2 represents ASU, its first and second derivatives which are used during backpropagation.

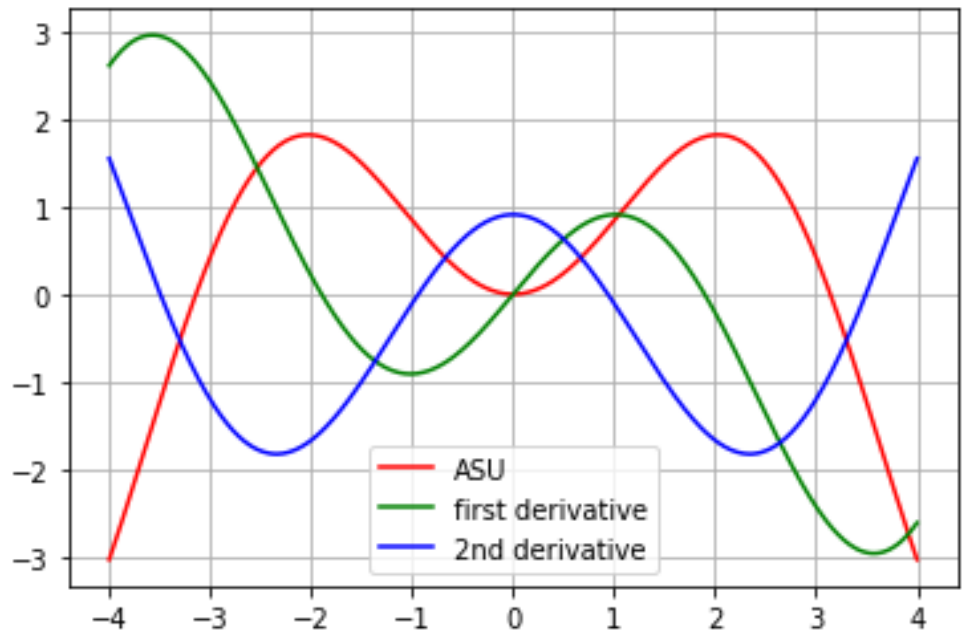

*Fig. 2  Plots of ASU along with its first and second derivatives*

ASU mimics the same oscillatory behavior for both strictly positive and strictly negative sides of domain and the oscillations tend to amplify for larger input values. The first derivative as shown by green curve in Fig. 2 behaves exactly opposite along both sides of zero and going farther along the domain, oscillations again tend to amplify. The second derivative also oscillates and magnify for higher inputs and behaves symmetrically along zero of domain.

**2.1.2 Pooling layer**

Pooling layers are basically used to downsize the feature maps in order to reduce computational cost. Max pooling has been in common practice which selects the most prominent feature value inside the pooling window. In this work, the first two convolutional layers of the network are set to a max pooling of size (2, 2) with stride 2. The task of feature extraction is being done hierarchically in the convolutional layers of the network. Fig. 3 and Fig. 4 represents the feature maps obtained from different layers of designed CNN architecture.

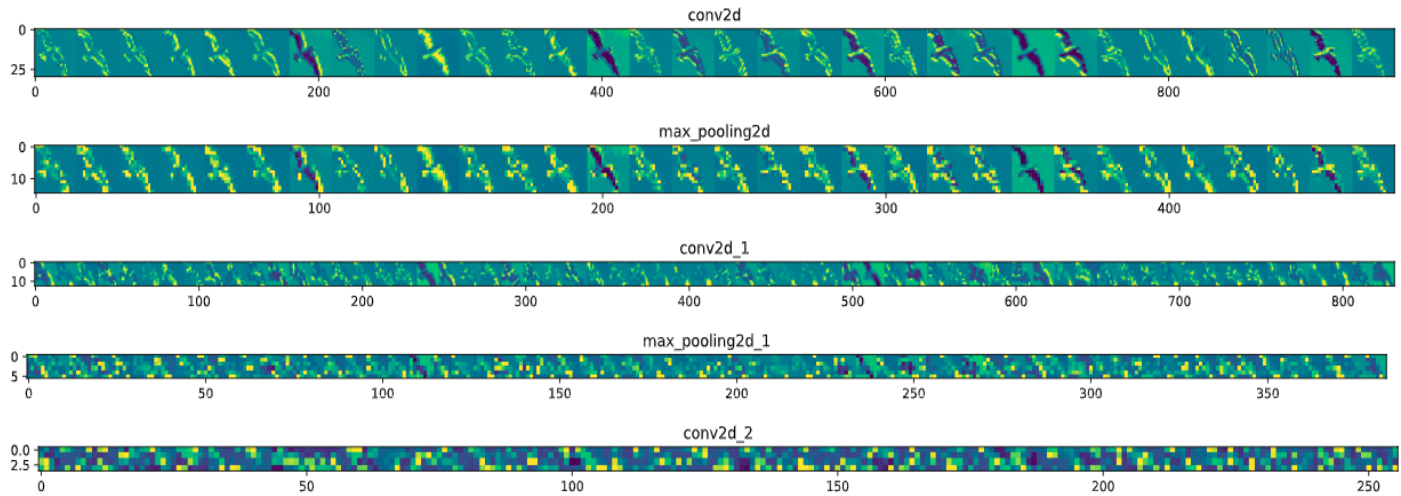

*Fig. 3 Feature maps' visualization from different layers of CNN when a bird image is fed to the visualization model*

The first convolutional layer i.e. conv2d applies 32 filters which detect major features like edges, horizontal or vertical patterns present in wings of the bird and its tail. Going deeper towards conv2d_1, the focus is to observe relatively complex patterns present in the input. In the $3^{rd}$ layer conv2d_2, the networks zoom more to get even the minor details of patterns or in other words it focuses on high level feature detection.

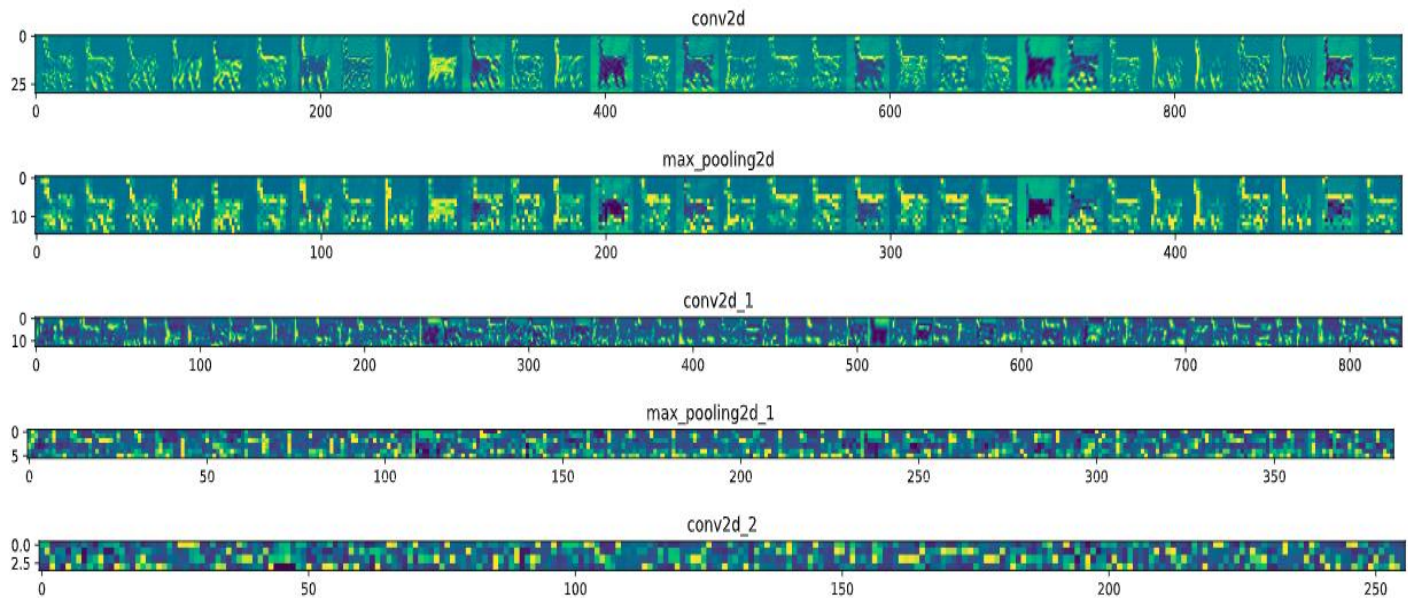

*Fig. 4 In case of a cat image, visualization of feature maps from different convolutional layers of network*

The same hierarchy of feature detection works for the input of a cat image. One can observe that the initial layers extracts low level features, focuses on general patterns present in the image such as presence of tail, legs, edges and curves of body. Going deeper towards conv2d_2 in the network the feature extractions become more complex and particular.

## 2.2 Dense layers

These are fully connected layers resembling multi-layer perceptron (MLP) models. Output from the last convolutional layer is flattened and send to the dense layers which work like a standard MLP network for classification. The designed network have two dense layers; one hidden layer with ASU as activation and other the output layer with softmax activation function.

## 3 Experimental setting

The designed CNN shown in Fig. 1 consists of three convolutional layers followed by a dense network with one hidden layer and an output layer of 10 neurons. Weights of network are initialized by Xavier uniform initializer. For training, the network utilizes Sparse Categorical Cross-entropy loss which is then optimized through Adam. The network is set to train for 20 epochs.

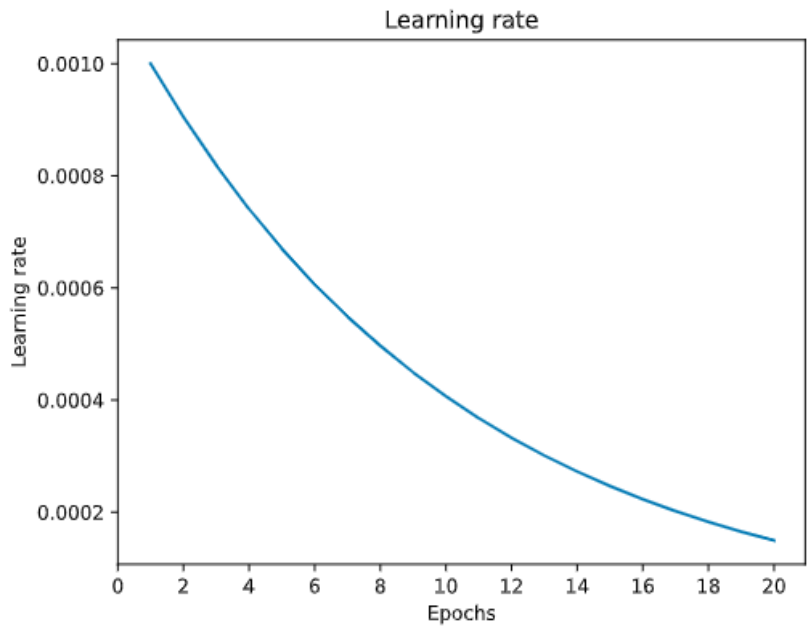

*Fig. 5* *Exponential decay of learning rate starting from 0.001 with a decay rate of 0.1 reaching to 0.00015 at $20^{th}$ epoch*

Initial learning rate is set to 0.001 with an exponential decay rate of 0.1 shown in Fig. 5. The detailed layer-structure of designed CNN architecture is presented by Fig. 6.

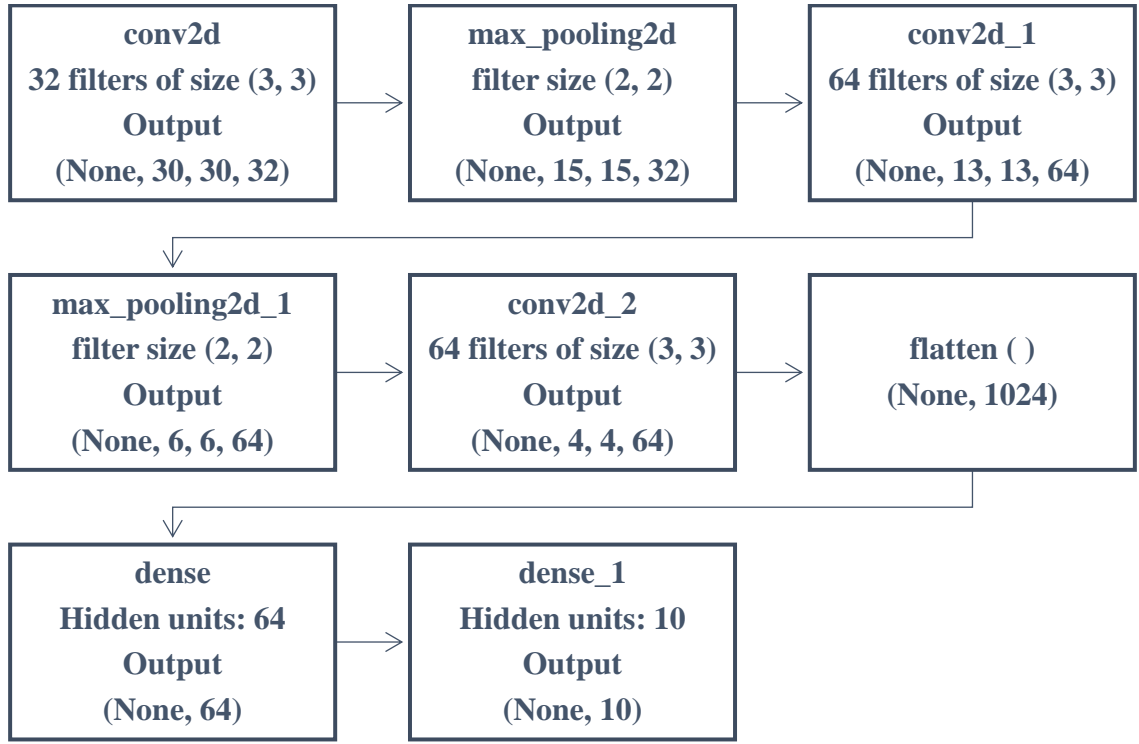

*Fig. 6 Experimental construction of CNN with layer names, filters and output dimensions of feature maps coming out from each convolutional layer*

**4 Results and discussion**

    The experimental results of the proposed network utilizing the ASU activation function for classification of the CIFAR-10 dataset are summarized in this section. The network was evaluated based on its performance in terms of training time, accuracy on training data, loss history during training, and accuracy on testing data. During a single run consisting of 20 epochs, the network completed the training process in 3 minutes and 2 seconds. Throughout this period, the network achieved an accuracy of 90.15% on the training data, indicating its capability to learn and classify the dataset effectively. Figure 7 displays the loss history of the network during the training process. It is evident from the graph that as the number of epochs increased, the training loss progressively decreased. This implies that the network improved its ability to minimize the discrepancy between predicted and actual values as it underwent more training iterations. Moreover, the network maintained a reasonably low validation loss, indicating its generalization capability on unseen data. Figure 8 presents the training and validation accuracy of the network. The graph illustrates the progression of both accuracies throughout the training process. It is noteworthy that the network consistently improved its accuracy on the training data, reaching a high value of 90.15%. Meanwhile, the validation accuracy also demonstrated a satisfactory performance, suggesting that the network could effectively generalize its knowledge to new instances. When evaluated on the testing data, the network achieved an accuracy of 70.15%. This indicates its ability to correctly classify previously unseen samples from the CIFAR-10 dataset, although with a relatively lower

accuracy compared to the training data. Further analysis and optimization can be performed to enhance the network's performance on the testing data and potentially increase its accuracy. In short, the proposed network utilizing the ASU activation function exhibited promising performance in the classification of the CIFAR-10 dataset. It demonstrated efficient learning during training, as depicted by the decreasing training loss and increasing training accuracy. Moreover, the network showed good generalization capabilities with a reasonable validation loss. Although the accuracy on the testing data was lower than on the training data, it still attained a respectable accuracy of 70.15%, implying its ability to classify unseen samples.

The network with ASU bore less computational cost as there is just one trancendental function and one multiplication computation is required as shown by (1). So, the training of network with ASU was much faster and the resultant accuracies are much appealing.

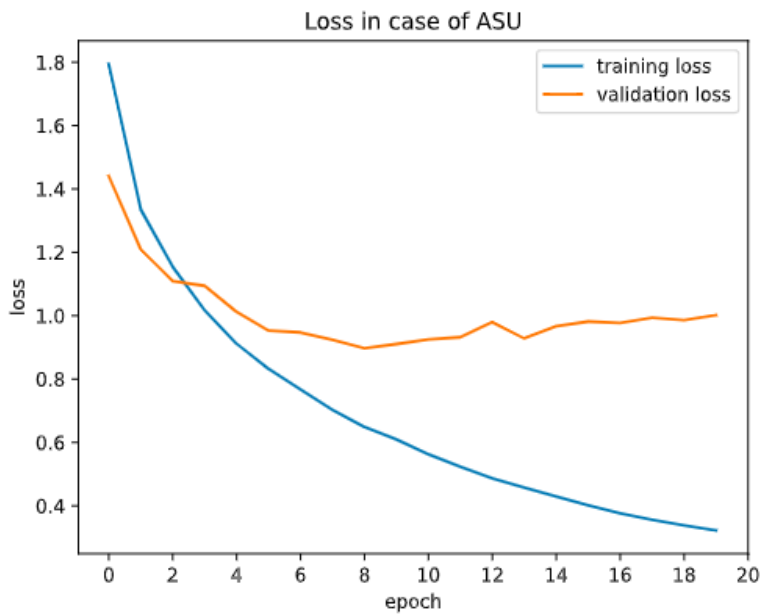

*Fig. 7  Training and validation loss history on CIFAR-10 in case of ASU activation for 20 epochs*

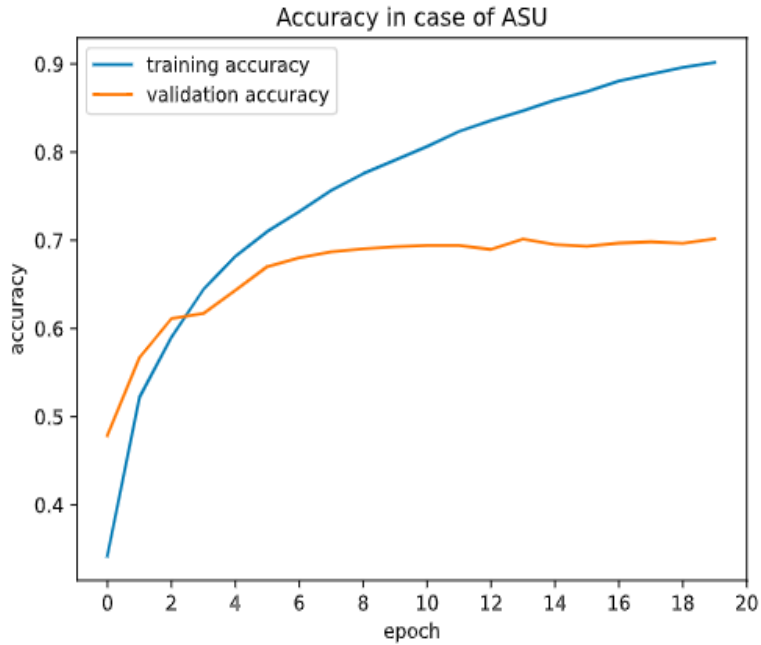

*Fig. 8* *Graphical illustration of training and validation accuracy of CNN on CIFAR-10*

**5 Conclusion**

      This article focused on evaluating the performance of ASU, an activation function, for image classification tasks. The designed CNN incorporating ASU throughout its layers was evaluated for CIFAR-10 and it demonstrated promising performance. The feature maps visualizations from all convolutional layers, training speed and computational feasibility of ASU are also authenticating the vigilance, prudence and effectivity of the proposed architecture. The results of the evaluation revealed promising performance, indicating that ASU has the potential to enhance the accuracy and effectiveness of CNN models for image classification tasks. As a future work this activation function can be tested for various other computer vision applications.